\documentclass{article}

\usepackage[preprint]{template}

\usepackage{natbib}
\setcitestyle{numbers,square}

\usepackage[utf8]{inputenc} 
\usepackage[T1]{fontenc}    
\usepackage{hyperref}       
\usepackage{url}            
\usepackage{booktabs}       
\usepackage{amsfonts}       
\usepackage{nicefrac}       
\usepackage{microtype}      
\usepackage{xcolor}         

\usepackage{graphicx}
\usepackage{commath}
\usepackage{subfig}
\usepackage{colortbl}

\title{Automatic Channel Pruning for Multi-Head Attention}

\author{%
  Eunho Lee, Youngbae Hwang \thanks{Corresponding author}\\
  Department of Intelligent Systems and Robotics\\
  Chungbuk National University\\
  Cheongju-si, 28644, Republic of Korea \\
  \texttt{\{ehlee, ybhwang\}@cbnu.ac.kr} \\
}

\begin{document}

\maketitle

\begin{abstract}
Despite the strong performance of Transformers, their quadratic computation complexity presents challenges in applying them to vision tasks.
Automatic pruning is one of effective methods for reducing computation complexity without heuristic approaches.
However, directly applying it to multi-head attention is not straightforward due to channel misalignment.
In this paper, we propose an automatic channel pruning method to take into account the multi-head attention mechanism.
First, we incorporate channel similarity-based weights into the pruning indicator to preserve more informative channels in each head.
Then, we adjust pruning indicator to enforce removal of channels in equal proportions across all heads, preventing the channel misalignment.
We also add a reweight module to compensate for information loss resulting from channel removal, and an effective initialization step for pruning indicator based on difference of attention between original structure and each channel.
Our proposed method can be used to not only original attention, but also linear attention, which is more efficient as linear complexity with respect to the number of tokens.
On ImageNet-1K, applying our pruning method to the FLattenTransformer, which includes both attention mechanisms, 
shows outperformed accuracy for several MACs compared with previous state-of-the-art efficient models and pruned methods.
Code will be available soon.
\end{abstract}

\section{Introduction}
\label{sec:intro}

Transformer has achieved remarkable success in various computer vision tasks based on attention mechanisms that effectively capture long-range dependencies.
The attention module generates an attention map by utilizing the query $Q\in \mathbb R^{N\times C}$ and key $K\in \mathbb R^{N\times C}$ to extract the relationships between tokens, and then projects the value $V\in \mathbb R^{N\times C}$ to obtain a feature map with global information.
It requires a computational cost of $\Omega(N^2C)$, which is quadratic with respect to the number of tokens $N$.
Despite providing excellent performance, the quadratic complexity with respect to $N$ poses significant challenges for deployment on mobile and edge devices.

Recent research attempts to mitigate this issue by designing efficient transformers.
Some approaches propose network architectures that limit the number of tokens~\cite{wang2021pvt, liu2021swin, hassani2023nat}.
By reducing the number of tokens, they cut down computation cost while trying to maintain the performance.
However, this results in limitations of the receptive field, constraining the capture of global dependencies.
Another approach proposes a new attention mechanism that can replace original attention~\cite{kitaev2020reformer, shaker2023swiftformer, han2023flatten}.
Linear attention is a method that maintains the ability to capture long-range dependencies while reducing computational complexity linearly.
They first approximate Softmax function by replacing it with a simple activation or a tailored function.
By changing the computation order from $(Q \cdot K^T) \cdot V$\footnote{Original attention mechanism} to $Q \cdot (K^T \cdot V)$\footnote{Linear attention mechanism}, they obtain an attention mechanism with computational complexity of $\Omega(NC^2)$, which is quadratic with respect to channels instead of tokens.

Another strategy to improve the efficiency of the network is network pruning.
It aims to achieve a lightweight network by removing redundant parts from the existing network while minimizing performance degradation.
Human knowledge-based analysis is conducted on each element such as gradient, Hessian distribution, and filter properties to measure the extent of redundancy~\cite{gao2023structural, yu2022unified, yang2023nvit}.
These methods require delicate tuning tailored to the network for optimal performance.
To tackle this, some approaches find the optimal network structure during the learning process.
Automatic pruning involves a hyper-network or pruning indicator with learnable parameters to the target network~\cite{xiao2019autoprune, li2020dhp, li2022pas}.
It finds the optimal configuration for each layer within a constrained budget through the learning process to achieve the best performance.
This eliminates the need for non-trivial hand-crafted designs tailored to specific networks.

\begin{figure}[t]
    \centering
    \includegraphics[width=\linewidth]{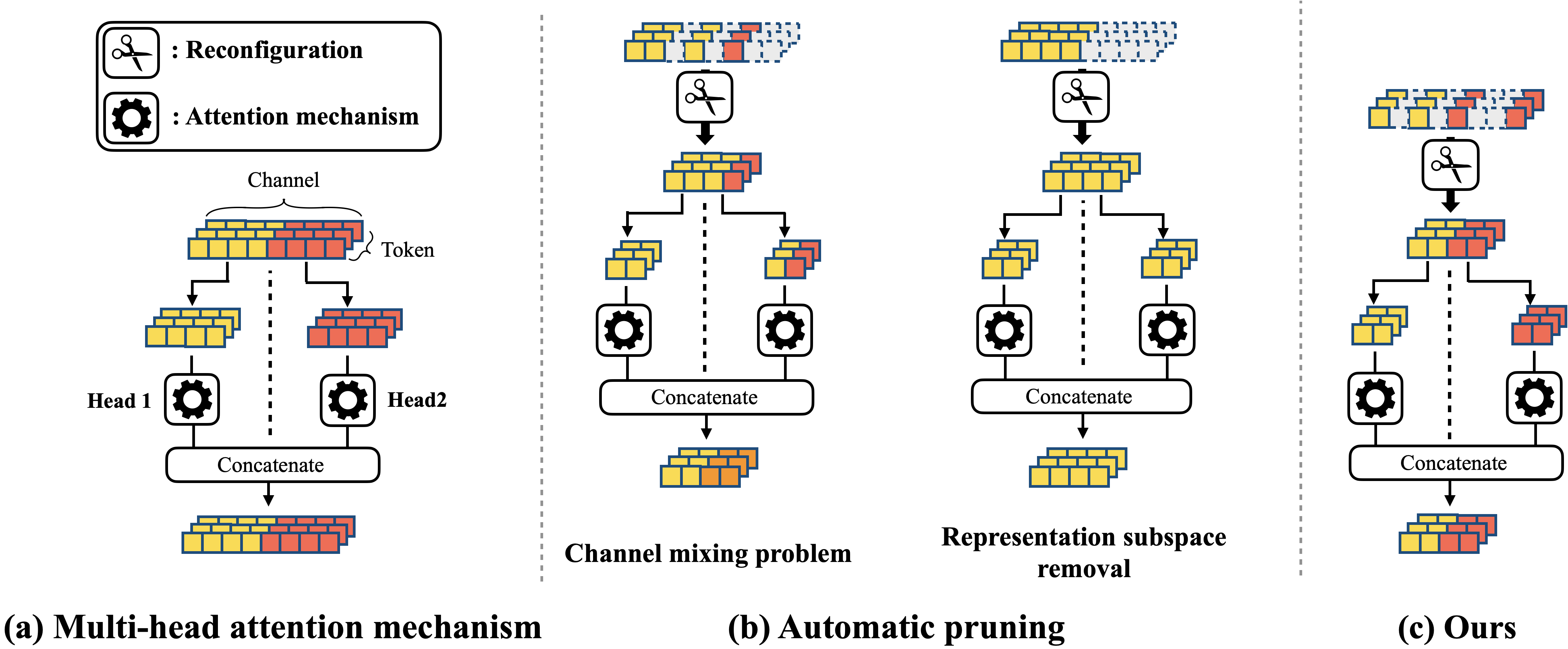}
    \caption{
    \textbf{Problems arising when multi-head is not considered.}
    (a) Multi-head attention is applied to each head, forming different representation subspaces.
    (b) If multi-head is not considered, reconfiguration leads to channel misalignment problems where representation subspaces get mixed, or subspaces are completely removed. It causes significantly reduction of the expression capacity.
    (c) Our method resolves these issues through an automatic pruning method by considering multi-head.
    }
    \label{fig:intro}
\end{figure}

Despite the strength of automatic pruning, applying it directly to Transformer is not trivial.
The original attention mechanism is structured with multiple heads, as shown in Fig.~\ref{fig:intro}~(a).
It enables the model to collectively attend to information from various representation subspaces at different positions~\cite{vaswani2017transformer}.
Fig.~\ref{fig:intro}(b) illustrates the issues when applying automatic pruning without considering multi-head attention.
If all channels of a specific head are removed, the corresponding representation subspace is eliminated, leading to limitations in the features that can be extracted.
As a result, there is a significant degradation in the expression capacity compared to the original network.
Pruning without considering multi-head attention leads to mixed channels from different representation subspaces, resulting in the extraction of features that are entirely different from those of the original network.
We refer to this issue as the \emph{channel misalignment problem.}

In this paper, we propose an automatic pruning method that takes into account the multi-head attention mechanism.
As in previous approaches~\cite{xiao2019autoprune, li2022pas, lee2024pruningfromscratch}, we apply a pruning indicator to estimate and remove redundant elements through learning.
During the training of the pruning indicator, we incorporate weights based on head-wise similarity to preserve more informative factors for each head.
This approach ensures that the representation subspaces remain similar to the original attention mechanism, minimizing the loss of expression capacity.
To prevent the channel misalignment problem that occurs during the reconfiguration process, as shown in Fig.~\ref{fig:intro}(b), we propose a pruning indicator adjustment process, which consists of head-wise ranking and rank-based average.
It resolves the channel misalignment problem by equalizing the pruning proportions for each head, as shown in Fig.~\ref{fig:intro}(c).
Our method utilizes similarity between each channel when training the pruning indicator, allowing removed channels to be represented by the combination of remaining channels.
Consequently, we compensate for information loss by adding a reweight module to adjust the scale of the remaining channels.
We present a method to initialize the pruning indicator for linear attention to prevent excessive channel removal.

Our method can effectively extend beyond original attention to include linear attention, without requiring tailored heuristics.
We apply it to FlattenTransformer~\cite{han2023flatten}, a model integrating both attention mechanisms, to discover the optimal model at the target compression rate.
Through comparisons with various manually designed models and pruned models on the ImageNet-1K benchmark~\cite{deng2009imagenet}, our approach demonstrates improved efficiency and superior accuracy for various MACs.
It is worthy note that compared with NViT~\cite{yang2023nvit}, a model obtained through structure pruning, our approach demonstrates approximately 0.3\% better performance at equivalent computational cost.
Even when compressing Flatten-Swin-T, our method exhibits superior performance by up to 0.25\%, despite being more efficient.

\section{Related Works}
\label{sec:related_works}
\subsection{Efficient Vision Transformer}
Transformer~\cite{vaswani2017transformer} exhibits exceptional performance in the NLP field due to their ability to effectively capture long-range dependencies.
The Vision Transformer (ViT) has successfully adapted these models for image classification, achieving outstanding performance~\cite{dosovitskiy2021vit}. 
However, the quadratic computational complexity of the original attention module in Transformers poses challenges for various vision applications.
Several methods address this concern by limiting the number of tokens.
Pyramid Vision Transformer (PvT)~\cite{wang2021pvt} progressively limits the number of tokens using spatial reduction attention, which controls the feature map size in the patch embedding.
Deformable Attention Transformer (DAT)~\cite{xia2022dat} selects the positions of keys and values in a data-dependent manner, considering only attentive tokens.
SwinTransformer~\cite{liu2021swin} divides the input into windows and performs original attention within these limited regions.
Neighborhood Attention Transformer (NAT)~\cite{hassani2023nat} localizes original attention by considering only the nearest neighboring pixels related to the query.
Other researches improve efficiency by integrating convolution operations into transformer models.
Convolutional vision Transformer (CvT)~\cite{wu2021cvt} uses convolutional projections instead of linear projections to control efficiency.
CMT~\cite{guo2022cmt} proposes a hybrid network of transformers and convolutions, achieving a better trade-off between accuracy and efficiency.

Another approach to achieve efficient vision transformers is to approximate the original attention with linear complexity operations.
EdgeNeXt~\cite{maaz2022edgenext} applies transposed attention along the channel dimension instead of the spatial dimension, achieving linear complexity with respect to tokens.
Reformer~\cite{kitaev2020reformer} replaces the dot-product operation with locality-sensitive hashing, reducing the computation to $\Omega(n \log n)$.
LinFormer~\cite{wang2020linformer} approximates the original attention matrix using low-rank matrix factorization, achieving linear complexity.
SwiftFormer~\cite{shaker2023swiftformer} demonstrates that key-value interaction can be replaced with a linear layer without performance degradation, effectively reducing computation.
A different line of research involves changing the order of original attention computations to achieve a complexity of $\Omega(NC^2)$.
This requires effectively replacing Softmax.
CosFormer~\cite{zhen2022cosformer} replaces Softmax with ReLU activation and cosine-based distance re-weighing.
SOftmax-Free Transformer (SOFT)~\cite{lu2021soft} uses a Gaussian kernel function to replace Softmax and achieves linear complexity through low-rank decomposition.
Castling-ViT~\cite{you2023castling} extracts spectral similarity between Q and K using a linear angular kernel.
FLattenTransformer~\cite{han2023flatten} reduces computation to linear complexity with focused linear attention, effectively maintaining the expressiveness of original attention.

\subsection{Network Pruning}
Pruning involves identifying and removing redundant components to reduce the network size.
Pruning has proven highly effective in original Convolutional Neural Networks (CNNs)~\cite{lin2020hrank, hou2022chex, gao2023structural, chen2023slats}.
Building on the strong performance of CNNs, there has been researches aimed at enhancing the efficiency of Vision Transformers through pruning.
One approach, token pruning, involves finding and removing unnecessary tokens, effectively reducing the computational load of the attention mechanism~\cite{bolya2022tokenmerging, wei2023tps, tang2023dynamictoken}. 
Another approach is to reduce the size of the Transformer model itself.
SViTE~\cite{chen2021svite} determines redundant components via sparse training and prunes accordingly.
Unified Visual Transformer Compression (UVC)~\cite{yu2022unified} reduces network size by combining pruning methods with various compression techniques.
Novel ViT (NViT)~\cite{yang2023nvit} enables global structural pruning based on a Hessian-based structural pruning criterion.
It uses only their heuristic criterion when pruning multi-heads. In contrast, the proposed method takes into account the representation space of each head, minimizing the information loss of the original model's attention mechanism.

To address these issues, there are automatic pruning methods that identify redundant components through training.
AutoPrune~\cite{xiao2019autoprune} determines the necessity of weights based on trainable auxiliary parameters.
MetaPruning~\cite{liu2019metapruning} involves training a meta-network, PruningNet, to find the optimal structure for a given target network.
DHP~\cite{li2020dhp} applies differentiable hypernetworks to determine the configuration of each channel in the backbone network.
Instead of using a hypernetwork, another approach involves a pruning module to each component, learning the importance and removing components~\cite{li2022pas, lee2024pruningfromscratch}.
VTP~\cite{zhu2021trainable} applies automatic pruning to Transformers by using trainable gate variables to identify and remove unnecessary components through learning.
Automatic pruning can identify the optimal network structure without heuristics, but directly applying them to Transformers is not trivial.
Unlike the traditional CNN structure, an automatic pruning method should take into account the multi-head structure in Transformers without losing expression capacity. 
In this paper, we propose a multi-head-aware automatic pruning method, which can be applied to both original attention and linear attention mechanisms.

\section{Method}
\label{sec:method}

\subsection{Multi-head Automatic Pruning}
\label{sub:multi-head}
\begin{figure}[t]
    \centering
    \includegraphics[width=\linewidth]{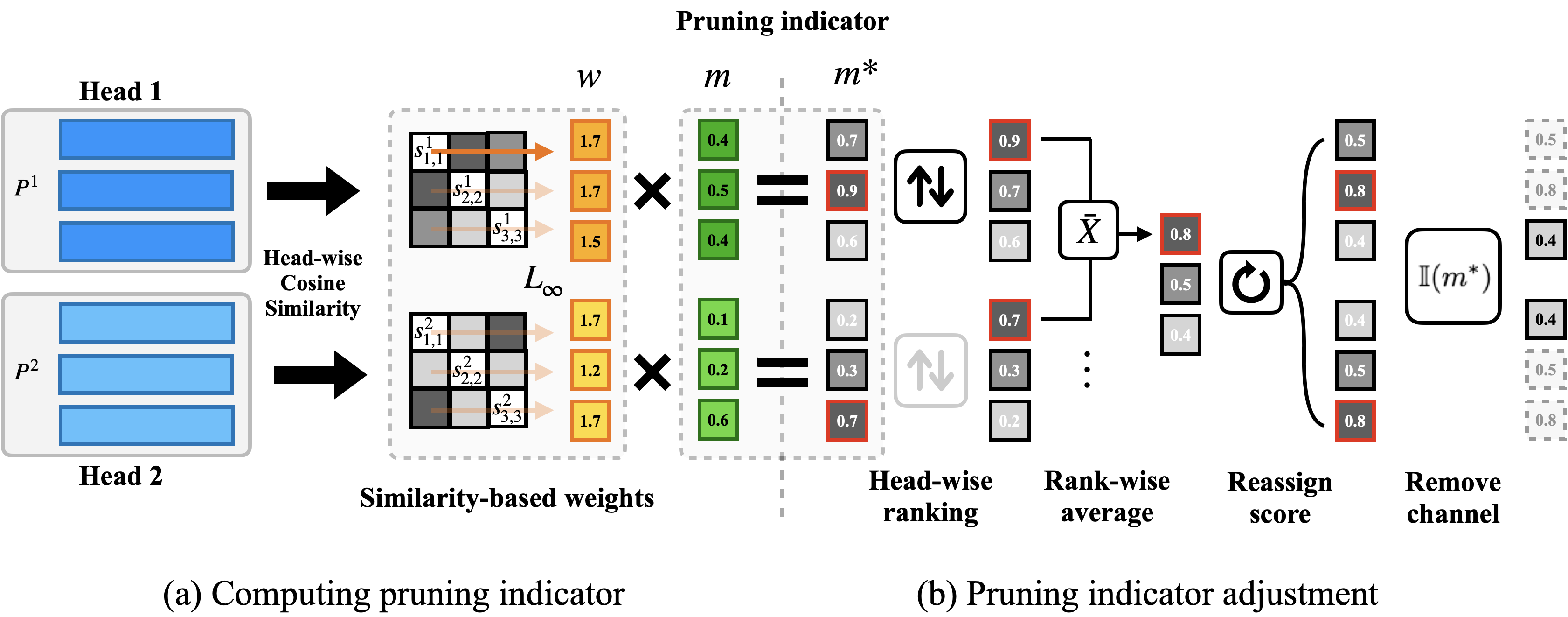}
    \caption{
    \textbf{Multi-head Automatic Pruning process.}
    To consider multi-head, we first undergo the (a) Computing pruning indicator process.
    We incorporate similarity-based weights into pruning indicators, enabling the pruning indicator to consider saliency channels in each head.
    (b) Through Pruning indicator adjustment, we share rank-wise pruning indicators for each head.
    It ensures equal channel removal across all heads, preventing the channel misalignment.
    }
    \label{fig:pruning_method}
\end{figure}

In this section, we propose an automatic pruning method for handling multi-head models.
It consists of two processes, computing pruning indicator and adjusting pruning indicator.
Fig.~\ref{fig:pruning_method}(a) illustrates a step of computing pruning indicator, where the importance of each channel is estimated to assign scores.
As in previous automatic pruning methods, a pruning indicator consisting of learnable parameters $m \in \mathbb R^{C_{out}}$ is employed to determine importance during training.
We assign weights to the pruning indicator based on head-wise similarity, considering more informative elements of the representation subspace.
First, the projection matrix $P\in \mathbb R^{C_{in} \times C_{out}}$ is divided into $P^1, P^2, ..., P^h \in \mathbb R^{C_{in}\times C_{h}}$  based on the number of heads $h$, where $C_{in}, C_{out}$ and $C_{h}$ are the number of input, output and head channels, respectively.
To compute the importance, we first compute the cosine similarity of the channels for each head.
\begin{equation}
    \centering
    S^k = \frac{P^k \cdot (P^k)^T}{\norm{P^k} \norm{P^k}} = 
    \begin{pmatrix}
        s^k_{1,1}    & \cdots    & s^k_{1,C_{h}} \\
        \vdots      & \ddots    & \vdots   \\
        s^k_{C_{h},1}    & \cdots    & s^k_{C_{h},C_{h}} \\
    \end{pmatrix}
    \label{eq:similarity}
\end{equation}
where $S^k$ is the similarity matrix obtained from the cosine similarity of $P^k$, where element $s^k_{i,j}$ represents the relationship between the $i$-th channel and the $j$-th channel in head $k$.
To incorporate this into the pruning indicator $m$, we calculate the similarity-based weights $w\in\mathbb R ^{C_{out}}$ as follows.
\begin{equation}
    \begin{split}
    w^k_i = & 1 + \lim_{p\to\infty} ( \sum_{n=1, n \neq i}^ {C_h} |s^k_{i,n}|^p)^{\frac{1}{p}}, \\
    w = & \text{Concat}[w^1, w^2, \cdots, w^h].
    \end{split}
    \label{eq:weight}
\end{equation}
For each row, we estimate similarity by applying the Chebyshev norm ($L_{\infty}$) to the remaining elements excluding itself.
Through this process, $w_i^k$ represents the similarity between the $i$-th channel and the other channels in head $k$.
If $w_i^k$ has a high score for a specific channel, it indicates the presence of other channels with high similarity, implying that it can be replaced by those channels.
By combining $w_1^k, w_2^k, \dots, w_{C_h}^k\in w^k$ using the concatenation operation Concat$(\cdot)$, we obtain the similarity-based weight $w$.
These values are weighted in the pruning indicator as follows.
\begin{equation}
    m^* = w \odot m.
\end{equation}
where $\odot$ represents the element-wise product.
Fig.~\ref{fig:pruning_method}(b) illustrates the method for adjusting pruning indicators to ensure equal distribution of pruned channels across all heads.
First, the rank for each head is computed according to $m^*$.
The score is calculated by taking the average of $m^*$ with the same rank in each head.
This average is then reassigned to the head positions with the same rank, enabling the adjustment of pruning indicators to share the same importance score across all heads.
Consequently, redundant channels are masked out by passing through the following indicator function:
\begin{equation}
    \mathbb{I}(m^*) =
\begin{cases}
    1,          & \text{if } m^* < \tau \\
    0,          & \text{otherwise.}
\end{cases}
\label{eq:binary_mask}
\end{equation}
Since the same rank across all heads holds the same pruning indicator, the removal of channels corresponding to each rank is simultaneously determined.
This allows training to ensure that pruning occurs at the same proportion across all heads, thereby preventing channel misalignment.

\subsection{Reweight Module}
\label{sub:reweight}
The removal of channels results in information loss, requiring a method to compensate for this loss to minimize performance degradation.
The proposed method learns indicators by considering the similarity of channels for each head, and subsequently removes channels based on these values.
The channels being removed can be represented by the weighted sum of the remaining channels, due to their high similarity with other channels.

For the reweighting method, a simple channel attention mechanism inspired by SENet~\cite{hu2018senet} is employed.
The input token $F\in\mathbb R^{N\times C}$ is compressed into a feature map $F_C \in \mathbb R^{1\times C}$ by taking the mean.
The compressed feature map is encoded into weights to compensate for each channel through a linear layer and $Tanh$ activation function.
This is multiplied with the query, allowing the passage of information from the removed channels to the remaining channels.
By applying this module, compensation can be provided for the removed channels, minimizing performance degradation.

\subsection{Pruning Indicator Initialization for Linear Attention}
\label{sub:init}
Unlike original attention in Transformer, linear attention with a computational complexity of $\Omega(NC^2)$, significantly improves in efficiency as channels are removed.
Even though the channels of linear attention are likely to be severely removed during the pruning process, it is non-trivial to properly determine an initial constant value to prevent it.
As a result, sophisticated initialization of the pruning indicator is required.
We introduce a data-driven method to solve the problem.
When the embedding vector is received as input, it is projected into query $Q \in \mathbb{R}^{N \times C}$, key $K \in \mathbb{R}^{N \times C}$, and value $V \in \mathbb{R}^{N \times C}$.
The relationship between each token $Q {K}^T$ can be expressed as a linear combination of each channel as shown below.
\begin{equation}
    QK^TV = (Q_1 K^T_1 + Q_2 K^T_2 + \cdots + Q_C K^T_C) V.
    \label{eq:init}
\end{equation}
The importance of each channel $Q_i K_i^T V$ can be determined by calculating the matrix distance from $QK^TV$ in Eq.~\ref{eq:init}.
By projecting $Q_i K_i^T$ onto $V$, we compare matrices in $\mathbb R^{N\times C}$ instead of $\mathbb R^{N\times N}$, significantly reducing computational complexity when $N\gg C$.
We use the difference in singular values to measure the matrix distance.
Let $\Sigma^{j}$ and $\Sigma^{j}_i$ be the singular values of the $QK^TV$ and the $Q_i K_i^T V$ in $j$-th image of training database, respectively.
The importance score of channel i is given by:
\begin{equation}
    T^{j}_i = |\Sigma^{j}_i - \Sigma^{j}_i|.
\end{equation}
The pruning indicator is initialized by accumulating the distance differences for each image and normalizing them between 0 and 1.
Using this initialization, we can retain a sufficient number of channels, ensuring the attention mechanism operates effectively.

\subsection{Training Automatic Pruning Method}
\label{sub:loss}
In the $l$-th layer of the Transformer, the pruning module ${m^*}^l$ is applied to the original model's weight $o^l\in \mathbb R^{C_{in}\times C_{out}}$ as follows:
\begin{equation}
    \tilde{o}_i^l=(\mathbb I ({m^*}^l_i)\odot o_i^l).   
\end{equation}
where $o^l_i$ and $\tilde{o}_i^l$ denote the weights of the $i$-th channel in the original and pruned layers, respectively.
$\mathbb I(\cdot)$ is the indicator function in Eq.~\ref{eq:binary_mask}.
Since $\mathbb I(\cdot)$ is a non-differentiable binary operation, the straight-through estimator (STE) is applied in back-propagation, allowing the pruning indicator to be learned by directly passing the gradient from $\mathbb I(m_i^l)$ to ${m^*}_i^l$.

The loss for the automatic pruning method is as follows:
\begin{equation}
\mathcal L=\mathcal L_{CE}(f(x), y) + \mathcal L_{MAC}(M_{prune}, M_{target}) 
\end{equation}
where $L_{CE}$ is the cross-entropy loss, comparing the model output $f(x)$ for input $x$ with the true label $y$.
MAC aware loss $\mathcal L_{MAC}$ is the Euclidean distance between the current pruned model's MAC, $M_{prune}$ and the target MAC, $M_{target}$.
To compute the current MAC, the following formula is used:
\begin{equation}
    M_{prune}=\sum_{l\in P}^lN\times \tilde{C}_{in} \times \tilde{C}_{out} + \sum_{l\in LA}^l (N\times \tilde{C}_{q} \times \tilde{C}_{k} + N\times \tilde{C}_{k} \times \tilde{C}_{v}) + \sum_{l \in OA}^l(N^2\times \tilde{C}_q + N^2 \times \tilde{C}_v)
\end{equation}
where $P$ denotes the projection matrix layer, and $\tilde{C}_{in}, \tilde{C}_{out},$ and $N$ represent the numbers of remaining input and output channels and tokens, respectively.
Unlike traditional CNNs, Transformers have additional computational costs due to the attention mechanism.
$LA$ and $OA$ represent linear attention and original attention, respectively, with $\tilde{C}_q , \tilde{C}_k,$ and $\tilde{C}_v$ indicating the remaining numbers of query, key, and value channels.
Considering the reconfiguration process, the pruning indicator $m^*$ for query and key projection is shared to ensure they are pruned at the same proportion.

\section{Experiments}
\label{sec:experiments}
To verify the proposed method, we apply our Automatic channel Pruning for Multi-head Attention (APMA) to the FLattenTransformer~\cite{han2023flatten}.
This network is composed of original attention blocks from the SwinTransformer~\cite{liu2021swin} and linear attention blocks.
Unlike previous pruning methods, the proposed approach can be applied to both types of attentions without specific heuristics.
APMA-B compresses the FLatten-Swin-S model to 6.2G, while APMA-S and APMA-T compress the FLatten-Swin-T model to 4.2G and 1.3G, respectively.
While other models follow the hybrid structure of FLattenTransformer, the smallest model, APMA-T, consists solely of linear attention to achieve a higher compression ratio.
We evaluate the performance of the compressed models on the classification task using the ImageNet-1K dataset~\cite{deng2009imagenet}.

To compress the network, we perform search and refine process.
The search process involves determining the optimal network through an automatic pruning module over 30 epochs.
We use the AdamW optimizer, starting with a learning rate of 5e-4 and decaying to 5e-6, with a weight decay set to 1e-6.
The batch size is 1024, and the training is conducted on 8 RTX A6000 GPUs.
The refine process aims to recover any information loss during the search phase.
Except for setting the weight decay, the experimental setup remains the same as in the search process.
APMA-B and APMA-S apply a weight decay of 0.05, while APMA-T uses a weight decay of 1e-6 to minimize additional sparsity.
In all experiments, the threshold $\tau$ for the indicator function is set to 0.5.

\subsection{Comparison with other methods}
\begin{table}[t]
\centering
\scriptsize
\begin{tabular}{l|ccc|l|ccc}
\hline
                                                             & Type & Acc. (\%) $\uparrow$ & MAC (G) $\downarrow$ & \multicolumn{1}{c|}{}                                        & Type                      & Acc. (\%) $\uparrow$      & MAC (G) $\downarrow$    \\ \hline
SViTE-B-40~\cite{chen2021svite}        & P    & 82.92                             & 11.7                              & T2T-ViT-14~\cite{yuan2021t2t}          & M                         & 81.7                                   & 6.1                                  \\
EViT-DeiT-B~\cite{liang2022evit}       & P    & 82.1                              & 11.6                              & CaiT-XS-24~\cite{touvron2021cait}      & M                         & 82.0                                   & 5.4                                  \\
AutoFormer-B~\cite{chen2021autoformer} & P    & 82.90                             & 11                                & AS-ViT-S~\cite{chen2021asvit}          & M                         & 81.2                                   & 5.3                                  \\
ConViT-S+~\cite{d2021convit}           & M    & 82.90                             & 10                                & TNT-S~\cite{han2021tnt}                & M                         & 81.5                                   & 5.2                                  \\
VTP-DeiT-B~\cite{zhu2021trainable}     & P    & 80.7                              & 10.0                              & DeiT-S~\cite{touvron2021deit}          & M                         & 81.2                                   & 4.6                                  \\
CaiT-S-24~\cite{touvron2021cait}       & M    & 83.5                              & 9.4                               & CvT-13~\cite{wu2021cvt}                & M                         & 81.6                                   & 4.5                                  \\
Swin-S~\cite{liu2021swin}              & M    & 83.00                             & 8.7                               & Swin-T~\cite{liu2021swin}              & M                         & 81.3                                   & 4.5                                  \\
FLatten-Swin-S~\cite{han2023flatten}   & M    & 83.50                             & 8.7                               & FLatten-Swin-T~\cite{han2023flatten}   & M                         & 82.1                                   & 4.5                                  \\
SPViT~\cite{he2024spvit}               & P    & 82.40                             & 8.4                               & GLiT-S~\cite{chen2021glit}             & M                         & 80.5                                   & 4.4                                  \\
CaiT-XS-36~\cite{touvron2021cait}      & M    & 82.90                             & 8.1                               & NViT-S~\cite{yang2023nvit}             & P                         & 82.19                                  & 4.2                                  \\ \cline{5-8} 
UVC-DeiT-B~\cite{yu2022unified}        & P    & 80.6                              & 8.0                               & \cellcolor[HTML]{FFFFC7}APMA-S                               & \cellcolor[HTML]{FFFFC7}P & \cellcolor[HTML]{FFFFC7}\textbf{82.35} & \cellcolor[HTML]{FFFFC7}\textbf{4.2} \\ \cline{5-8} 
STEP-Swin-S~\cite{li2021step}          & P    & 79.6                              & 6.3                               & AutoFormer-T~\cite{chen2021autoformer} & P                         & 75.70                                  & 1.3                                  \\
NViT-H~\cite{yang2023nvit}             & P    & 82.95                             & 6.2                               & NViT-T~\cite{yang2023nvit}             & P                         & 76.21                                  & 1.3                                  \\ \hline
\rowcolor[HTML]{FFFFC7} 
APMA-B                                                       & P    & \textbf{83.19}                    & \textbf{6.2}                      & APMA-T                                                       & P                         & \textbf{76.53}                         & \textbf{1.3}                                  \\ \hline
\end{tabular}
\caption{
\textbf{ImageNet-1K results} of various efficient models and our proposed method.
Our approach demonstrates higher performance at lower MAC than both manually designed networks (M) and efficient networks using pruning methods (P).
}
\label{tab:comparison}
\vspace{-0.5cm}
\end{table}

\begin{table}[t]
\centering
\small
\begin{tabular}{l|cccc}
\hline
                                     & Acc. (\%) $\uparrow$       & MAC (G) $\downarrow$ & Throughput $\uparrow$ & Speed up               \\ \hline
FLatten-Swin-S \cite{han2023flatten} & \multicolumn{1}{c}{83.5}   & 8.7                  & 492                   & \multicolumn{1}{c}{x1} \\
APMA-B                               & 83.19                      & 6.2                  & 663                   & x 1.4                  \\ \hline
FLatten-Swin-T                       & \multicolumn{1}{c}{82.1}   & 4.5                  & 805                   & x 1.6                  \\
APMA-S                               & 82.35                      & 4.2                  & 921                   & x 1.9                  \\
APMA-T                               & 76.53                      & 1.3                  & 1221                  & x 2.5                  \\ \hline
\end{tabular}
\caption{
\textbf{Comparison of Throughput} between FlattenTransformer and our proposed method.
All measurements are conducted under the same computational environment on RTX A6000.
Applying our proposed method demonstrates effectively improved inference speed.
Note that APMA-S achieves better throughput compared to Flatten-Swin-T, while also showing better performance.
}
\label{tab:speed}
\vspace{-0.5cm}
\end{table}
In Table~\ref{tab:comparison}, we compare the networks compressed using our proposed method with other manually designed models (M) and pruned networks (P).
Compared to manually designed models (M), our method demonstrates higher performance with lower computational cost.
For instance, when comparing APMA-B with Swin-S, our method achieves improved performance with approximately 30\% less computation.
Notably, APMA-S is more efficient than both Swin-T and DeiT-S, showing a significant performance improvement of about 1.15\%.

We compare our compressed networks with other pruned networks (P).
Our methods are shown to be the most efficient while maintaining high performance, surpassing the previous state-of-the-art NViT on all MACs.
APMA-B significantly reduces MACs while achieving superior accuracy compared to other pruning methods.
Particularly, the model composed solely of linear attention, APMA-T, demonstrates a 0.32\% higher performance at the same computational cost compared to other compressed networks. 
Other compressed networks are based on original attention with a computational complexity of $\Omega(N^2C)$.
In contrast, our proposed method compresses linear attention-based networks with a computational complexity of $\Omega(NC^2)$.
Consequently, even with less channel pruning, we achieve a significantly more efficient reduction in computational cost, offering greater expressive capacity at the same computational cost.

We also compare the performance of FLattenTransformer with our proposed method applied.
APMA-B compresses the FLatten-Swin-S model, achieving approximately 30\% less computational cost with only a 0.31\% performance drop.
However, as shown in Table~\ref{tab:speed}, there is a substantial throughput improvement of around 40\%.
APMA-S demonstrates a 0.25\% performance improvement over the FLatten-Swin-T model, despite having lower computational cost and faster computation speed.
This indicates the effectiveness of our method in finding the optimal model.
APMA-T, the fastest network, shows a 2.5x speed improvement over the FLatten-Swin-S model.

\subsection{Ablation Study}
\label{sub:ablation}
\begin{figure}[t]
    \centering
    \includegraphics[width=0.75\linewidth]{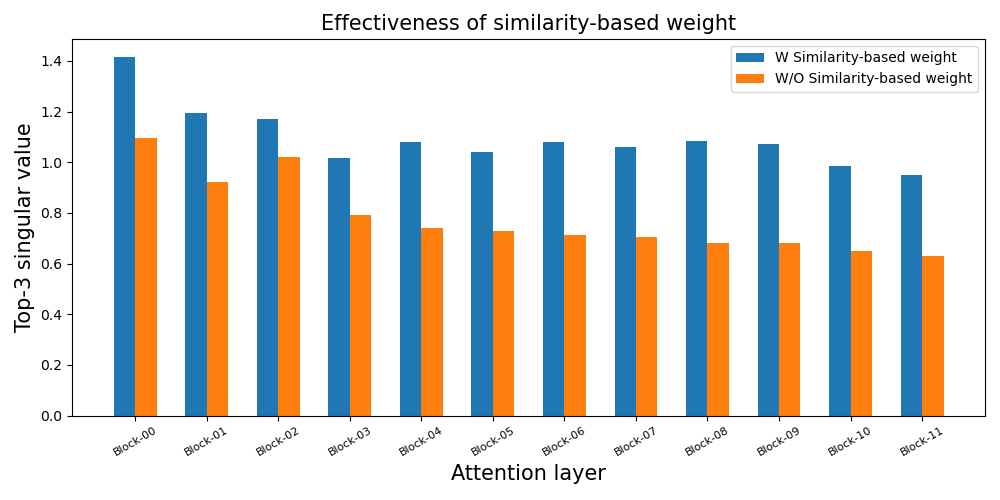}
    \caption{
    \textbf{Top-3 singular value norms} for each attention layer of the pruned model.
    When using Similarity-based weights, it demonstrates larger singular value norms.
    It indicates that when employing the proposed method, salient channels can be effectively retained.
    }
    \label{fig:similarity}
\end{figure}
\begin{table}[t]
\centering
\small
\begin{tabular}{l|cc}
\hline
                                 & \multicolumn{1}{c}{Acc. (\%) $\uparrow$} & \multicolumn{1}{c}{MAC (G) $\downarrow$} \\ \hline
Original                         & 76.53                                    & 1.3                                      \\ \hline
(-) Reweight module              & 76.36 (-0.17)                            & 1.3                                      \\
(-) Similarity-based weight      & 75.90 (-0.63)                            & 1.3                                      \\
(-) Multi-head automatic pruning & 72.78 (-3.75)                            & 1.3                                      \\
(-) Pruning indicator initialize & 71.04 (-5.49)                            & 1.3                                      \\ \hline
\end{tabular}
\caption{
\textbf{Evaluation of the effectiveness of proposed methods.}
Significant performance degradation is observed when each module and method is removed.
}
\label{tab:ablation}
\end{table}
In this section, we estimates the effectiveness of each module and method proposed by removing them individually and comparing the performance.
All experiments are conducted on APMA-T, and the experimental setup is consistently configured.
The Similarity-based weight and Multi-head automatic pruning are proposed in Sec.~\ref{sub:multi-head}, Reweight module in Sec.~\ref{sub:reweight}, and Pruning indicator initialization in Sec.~\ref{sub:init}.
\paragraph{Similarity-based Weight}
To demonstrate the effectiveness of the similarity-based weight method, we examine the singular value norms of each attention layer in the model immediately after the search process, as shown in Fig.~\ref{fig:similarity}.
Due to the varying number of channels in the models with and without the similarity-based weight, we calculate the norm of the top 3 singular values for each head.
Fig.~\ref{fig:similarity} shows that the singular value norms are higher across all layers when the similarity-based weight is used.
This indicates that the model possesses eigenvectors with higher importance, which means it can inherently extract more distinct features.
Consequently, as shown in Table~\ref{tab:ablation}, the performance difference between the model using the proposed similarity-based weight and the model without it is a significant 0.63\%.

\begin{figure}[t]
    \centering
    \includegraphics[width=0.8\linewidth]{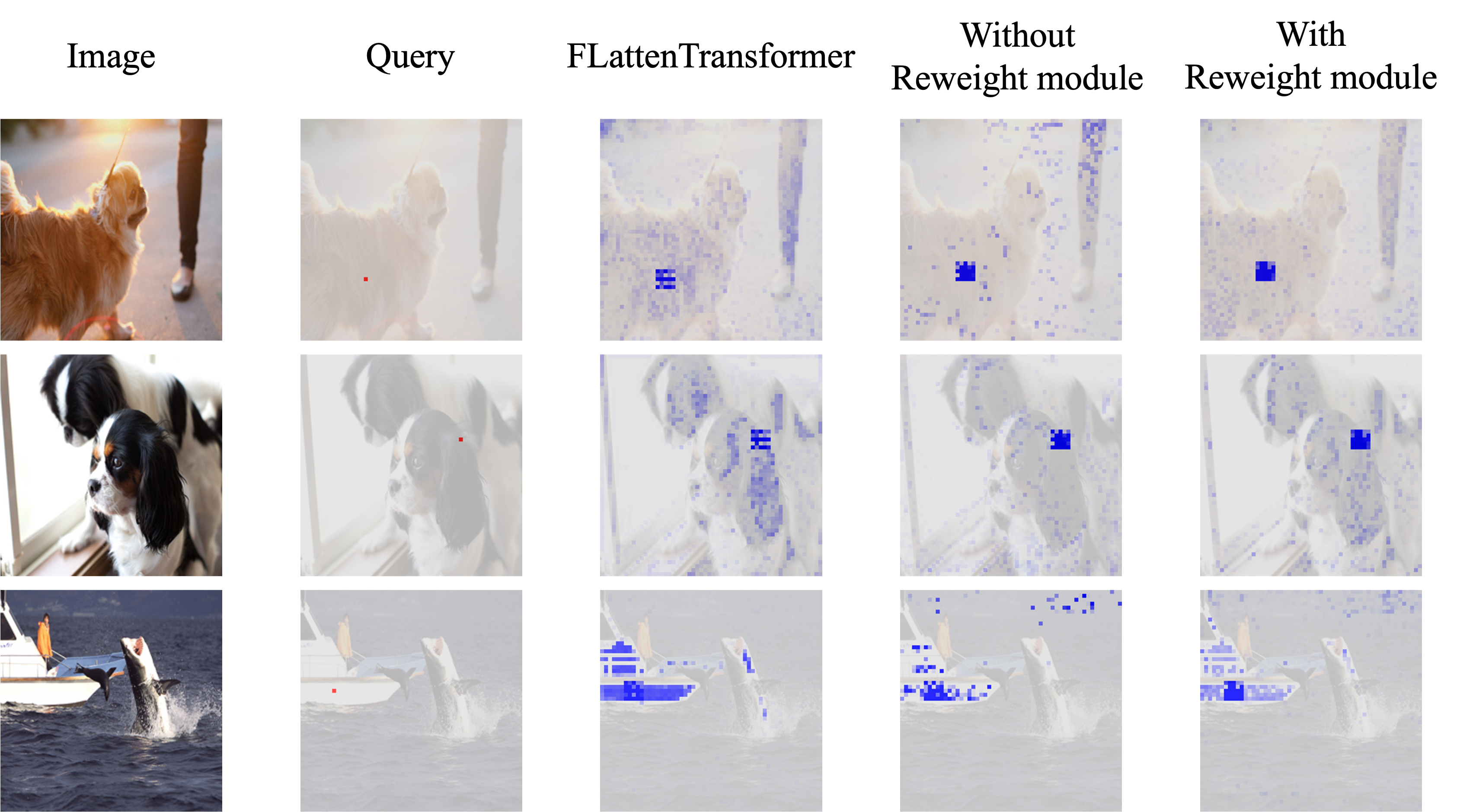}
    \caption{
    \textbf{The effect of the reweight module.}
    The red block represents the query token, while the blue blocks depict the relationship between this query token and other tokens.
    It demonstrates that the reweight module can compensate for information loss caused by pruning, directing attention to more relevant tokens.
    }
    \label{fig:reweight_exp}
\end{figure}
\paragraph{Reweight Module}
We apply a reweight module to compensate for the information loss that occurs when channels are removed.
As shown in Table~\ref{tab:ablation}, when the module is removed in the proposed automatic pruning method, there is a decrease in accuracy of 0.17 without any change in MAC.
Fig.~\ref{fig:reweight_exp} illustrates the attention for tokens with or without applying the module.
When applying the reweight module, the relationship with relevant object tokens strengthens, resembling the original model, while the relationship with background tokens weakens.
The result demonstrates that the reweight module compensates for information loss, allowing relevant tokens to receive better attention.

\begin{figure}[t]
    \centering
    \includegraphics[width=\linewidth]{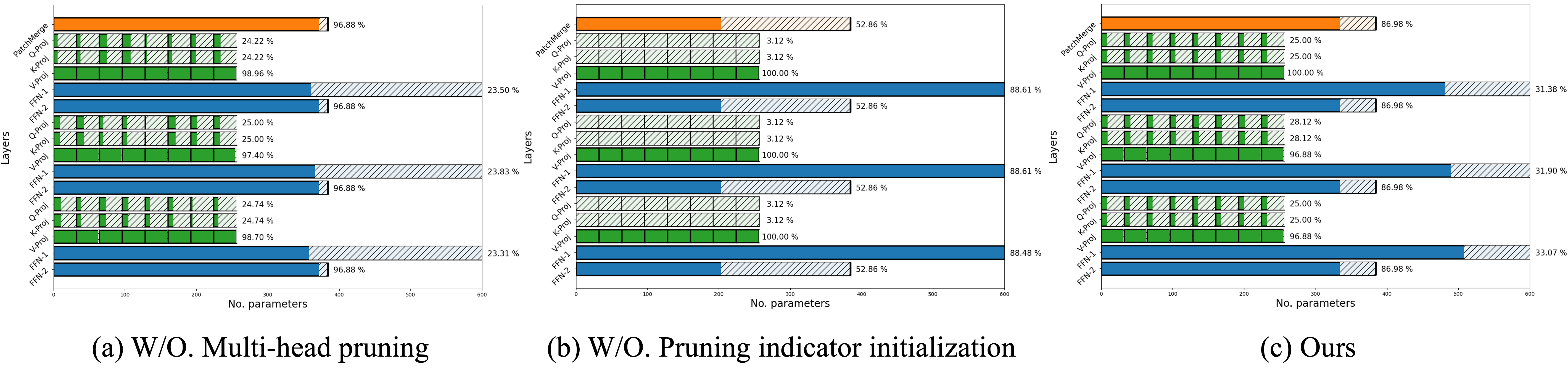}
    \caption{
    \textbf{Pruned model structure after reconfiguration.}
    (a) Without applying multi-head pruning, the pruning ratio for each head is inconsistent, leading to a channel misalignment problem. 
    (b) When pruning indicator initialization for linear attention is not performed, the attention mechanism does not function properly. 
    (c) Our proposed method ensures consistent pruning ratios across heads to resolve the channel misalignment problem and demonstrates effective pruning of each module in appropriate proportions.
    Zoom in for details.
    }
    \label{fig:structure_exp}
\end{figure}
\paragraph{Multi-head Automatic Pruning}
In contrast to conventional automatic pruning methods, our proposed approach can remove channels considering the multi-head structure.
Without considering multi-heads, the accuracy drops by 3.75, indicating a significant performance degradation, as shown in Table~\ref{tab:ablation}.
In Fig.~\ref{fig:structure_exp}(a), the channels for query, key, and value vary across each head.
During the reconstruction process in the network, where redundant channels are actually removed for real acceleration, channel misalignment occurs across heads.
On the other hand, applying our proposed method, as shown in Fig.~\ref{fig:structure_exp}(c), guarantees that channels are removed at the same proportion across heads.
In Fig.~\ref{fig:structure_exp}(a), the almost complete removal of channels in specific heads leads to restricted representation space and severe performance degradation.
In contrast, our proposed method ensures an equal distribution of channels per head, thereby preserving the network's expression capacity.

\paragraph{Pruning Indicator Initialize for Linear Attention}
Fig.~\ref{fig:structure_exp}(b) shows the model structure obtained from the search process when the pruning indicator for linear attention is not initialized.
It illustrates that channels of the projection matrix for the query and key are severely pruned, whereas the feed-forward network is not pruned.
This imbalance leads to improper operation of the attention mechanism.
In contrast, Fig.~\ref{fig:structure_exp}(c) demonstrates how our method alleviates such issues.
Table~\ref{tab:ablation} presents a significant performance degradation of 5.49\% when the pruning indicator is not initialized.
This shows the effectiveness of our proposed initialization method.

\section{Conclusion}
\label{sec:conclusion}
In this paper, we introduce an automatic pruning method tailored for multi-head attention mechanisms.
Leveraging similarity-based weights and a pruning indicator adjustment process, our approach handles multi-head attention, considering the representation subspace effectively.
Moreover, the integration of a reweight module compensates for the information loss incurred by pruned channels.
By initializing the pruning indicator for linear attention in a data-driven manner, our method facilitates the discovery of optimal models.
Comparative analysis against manually designed models and various pruning techniques demonstrates the superior efficiency and performance trade-off of our proposed approach.
In this paper, we only focus on experiments on the classification task.
Future work will extend our method to various tasks such as object detection and semantic segmentation.

\bibliographystyle{plain}
\bibliography{references}

\end{document}